\def\year{2022}\relax
\documentclass[letterpaper]{article} 
\usepackage{aaai22}  
\usepackage{times}  
\usepackage{helvet}  
\usepackage{courier}  
\usepackage[hyphens]{url}  
\usepackage{graphicx} 
\usepackage{natbib}  
\usepackage{caption} 
\DeclareCaptionStyle{ruled}{labelfont=normalfont,labelsep=colon,strut=off} 
\usepackage{algorithm}
\usepackage{algorithmic}

\usepackage{amsmath,amssymb} 
\usepackage{amsthm}
\usepackage{bbm}

\usepackage{multirow}
\usepackage{booktabs, microtype}
\usepackage{multicol}

\usepackage{hyperref}

\usepackage{newfloat}
\usepackage{listings}
\floatstyle{ruled}
\newfloat{listing}{tb}{lst}{}
\floatname{listing}{Listing}
\nocopyright
\title{Self-Supervised Contrastive Representation Learning for 3D Mesh Segmentation}
\author{
    Ayaan Haque\thanks{This work was conducted during an internship with Samsung SDS Research America. A. Haque is currently affiliated with UC Berkeley.},
    Hankyu Moon,
    Heng Hao,
    Sima Didari,
    Jae Oh Woo,
    Patrick Bangert
}
\affiliations{
    Samsung SDS Research America, CA, USA\\
    ayaanzhaque@berkeley.edu
}

\begin{document}

\maketitle

\begin{abstract}

3D deep learning is a growing field of interest due to the vast amount of information stored in 3D formats. Triangular meshes are an efficient representation for irregular, non-uniform 3D objects. However, meshes are often challenging to annotate due to their high geometrical complexity. Specifically, creating segmentation masks for meshes is tedious and time-consuming. Therefore, it is desirable to train segmentation networks with limited-labeled data. Self-supervised learning (SSL), a form of unsupervised representation learning, is a growing alternative to fully-supervised learning which can decrease the burden of supervision for training. We propose SSL-MeshCNN, a self-supervised contrastive learning method for pre-training CNNs for mesh segmentation. We take inspiration from traditional contrastive learning frameworks to design a novel contrastive learning algorithm specifically for meshes. Our preliminary experiments show promising results in reducing the heavy labeled data requirement needed for mesh segmentation by at least 33\%.

\end{abstract}

\section{Introduction}

\begin{figure*}
  \includegraphics[width=\linewidth, clip]{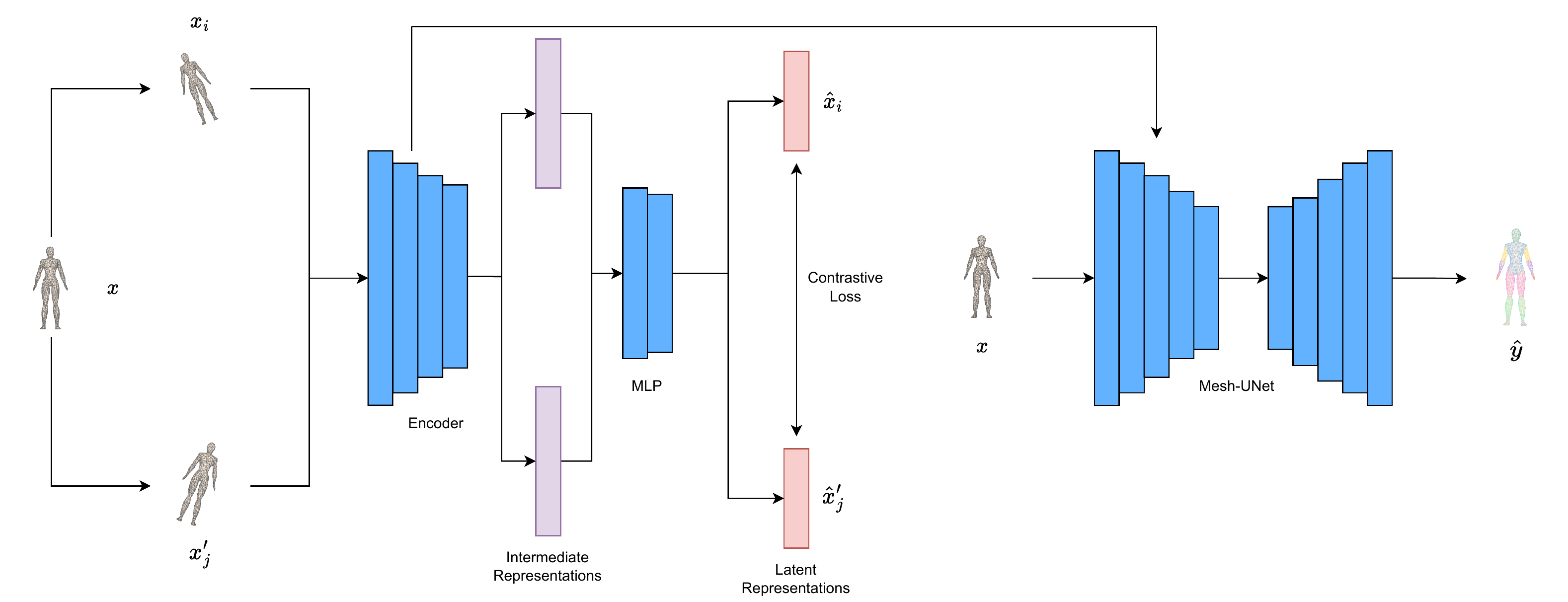}
  \caption{We pre-train a MeshCNN encoder using contrastive learning. The network learns to maximize agreement of representations between positive pairs while minimizing those of negative pairs. We then transfer the pre-trained encoder (shown with an arrow) to a Mesh-UNet to perform downstream segmentation.}
  \label{fig:algorithm}
\end{figure*}


3D polygonal meshes are a popular 3D data representation as they are efficient at representing objects with irregular and non-uniform surfaces. The polygonal faces of a mesh allow distinct surfaces to be represented through geodesic separation. Furthermore, meshes have improved efficiency over other representations such as point cloud because its faces can adapt to sharp features and flat regions.

Unfortunately, 3D data is challenging to analyze with CNNs, as traditional CNNs are designed to operate on the uniform, grid-like structure of an image. Various efforts have been made to process other forms of 3D and geometric data \cite{bronstein2017geometric, guo2020deep, he2021deep}. The pioneering method for working directly on meshes is MeshCNN \cite{hanocka2019meshcnn}. MeshCNN rethinks the convolution and pooling operations to take advantage of the non-uniformity and irregularity of the mesh, thus outperforming previous methods on various tasks such as segmentation which operate indirectly on meshes or on other data formats such as point cloud.

Traditionally, to train deep learning networks, large fully-supervised datasets are required. However, labeling meshes, and especially for segmentation, is challenging due to compute, cost, and time constraints. For example, when attempting to create groundtruth segmentations for medical meshes, an expert must spend extensive time to create precise and accurate labels. Due to these constraints, it may only be feasible to label a portion of the dataset. Therefore, it is desirable to train without large quantities of labeled data or build datasets where only part of the examples must be labeled. This problem is known as the ``limited-labeled data problem," or the ''small data problem."

Unsupervised representation learning, self-supervised learning (SSL), and contrastive learning are a family of training frameworks which have been successfully applied to addressing the small data problem. Unsupervised representation learning enables a network to learn strong visual representations without any supervision. Various forms include deep clustering \cite{huang2016unsupervised, caron2019unsupervised, zhan2020online} or contrastive learning. These methods can be categorized as generative or discriminative and have been successful in the 2D domain. In this work, we focus on discriminative methods for 3D mesh data.

Contrastive learning and SSL are forms of unsupervised representation learning. In SSL, a supervisory signal is produced synthetically from the unlabeled data itself. These labels are used to formulate a pre-text task, and the network is trained on this task to learn efficient representations. In contrastive learning, a network is trained to sort representations in the latent space based on their similarity to one another. There have been many successful recent works on contrastive learning \cite{hadsell2006dimensionality, dosovitskiy2014discriminative, oord2018representation, he2020momentum, caron2020unsupervised, moon2021patchnet, bangert2022medical, chen2021empirical}. Two competing standout methods, which many works have built upon, are SimCLR\cite{chen2020simple} and MoCo \cite{he2020momentum}. SimCLR \cite{chen2020simple} builds upon traditional contrastive learning, where positive and negative pairs are sorted, by simplifying the process, requiring no architectural or loss adjustments. MoCo, or momentum contrastive learning, builds dynamic dictionaries to perform dictionary look-up, where an encoded query should be similar to its matching key and dissimilar to others. 

In our work, we use contrastive learning as pre-training. The network learns efficient representations of all training samples (labeled or not), enabling it to perform strongly on the downstream task when trained only with labeled examples. While SimCLR is not currently state-of-the-art, our work aims to demonstrate that contrastive learning can be an effective combination for mesh segmentation. Accordingly, we adopt to use of the baseline SimCLR method.

Unsupervised representation learning has been applied to other formats of 3D data, primarily point cloud (see \citet{xiao2022unsupervised} for a survey). However, there have been limited works which apply unsupervised representation learning to meshes \cite{zhou2020unsupervised, wang2021nemo, zimmermann2021contrastive}. \citet{zhou2020unsupervised} learns to disentangle mesh representations using self and cross-consistency constraint for shape and pose estimation. \citet{wang2021nemo} learns a generative model for neural feature activations to perform pose estimation. \citet{zimmermann2021contrastive} proposes a MoCo-inspired contrastive learning method to perform hand shape estimation.

However, unsupervised representation learning methods for segmentation as well as SimCLR-based contrastive learning methods for mesh analysis remain under explored. In this paper, we introduce self-supervised MeshCNN, or SSL-MeshCNN, a novel mesh-specialized contrastive learning method to perform downstream segmentation with limited-labeled data.

\section{Methods}


A triangular mesh can accurately represent surfaces and topology of an object. A mesh is described by its vertex set $V\subset \mathbb{R}^3$, edge set $E$, and face set $F$. A face represents the triangle connectivity between vertices and is formalized as a tuple $\langle v_1, v_2, v_3 \rangle\in F$ for $v_1, v_2, v_3\in V$. Edge $E$ is a collection of the connection between two vertices, e.g., $\langle v_1, v_2\rangle\in E$ for $v_1, v_2\in V$ and is used to encode adjacency between two faces. Applying standard image-based CNNs to a triangular mesh is infeasible because meshes are irregular and non-uniform. MeshCNN \cite{hanocka2019meshcnn} was proposed to train directly on meshes by creating mesh-specific convolutional and pooling layers. Each edge in the mesh is used to create a 5-dimensional input feature set. This 5-dimensional edge feature consists of the dihedral angle, two inner angles and two edge-length ratios for each face. \citet{hanocka2019meshcnn} also proposed a Mesh-UNet which uses mesh unpooling layers in the decoder to perform segmentation. We use Mesh-UNet as the backbone segmentation network for our method.

Our overall training procedure involves two steps: contrastive learning pre-training using an entire dataset, followed by downstream segmentation using only samples with corresponding labels. Figure \ref{fig:algorithm} illustrates our contrastive learning framework. We can formulate our problem using dataset $\mathcal{D}_u$, which is the set of all meshes $x$. Not all samples $x$ in $\mathcal{D}_u$ have corresponding labels $y$. $\mathcal{D}_l$ is the subset of $\mathcal{D}_u$ containing only meshes $x$ that have corresponding labels $y$.

Contrastive learning learns efficient representations by maximizing agreement between two augmented versions of the same input in the latent space. For each input $x$ in minibatch of size $M_1$, using a stochastic augmentation function $\text{Aug}\left(\cdot\right)$, we generate two uniquely augmented meshes $x_i=\text{Aug}(x)$ and $x_j'=\text{Aug}(x)'$, which together form a positive pair ($x_i$, $x_j'$). This means each batch now has $2M_1$ total samples and $2(M_1-1)$ negative samples. For contrastive learning, our network architecture uses a MeshCNN encoder followed by a nonlinear two-layer multilayer perceptron (MLP) head. SimCLR utilizes a single-layer head, but rather we propose to use a two-layer head. Since the mesh representation has higher dimensionality than images, using two layers more appropriately reduces the mesh dimensionality, which is important for latent space sorting. We feed the batch into this architecture to receive the latent representations $\hat{x}_{i}$ and $\hat{x}_{j}'$. Contrastive loss is used to maximize agreement between representations of positive pairs and minimize agreement between representations of negative pairs. The goal is for the encoder to learn valuable representations, while the MLP only allows for better sorting of the representations in the latent space. Therefore, after pre-training, the MLP is discarded and only the pre-trained encoder is saved.

To perform SimCLR-based contrastive learning, a strong augmentation policy must be utilized in order to create effective positive and negative pairs. SimCLR originally was intended for images, and therefore uses three image-based augmentations. However, augmentations for meshes are significantly different. This is primarily because our input features are similarity-invariant, meaning augmentations such as isotropic scaling, rotation, and translation will result in no change to the mesh \cite{hanocka2019meshcnn}. Thus, we redesign this augmentation policy uniquely for meshes by carefully selecting our augmentations. We define a stochastic augmentation function $\text{Aug}\left(\cdot\right)$ which randomly applies a series of three augmentations: anisotropic scaling, vertex shifting, and edge flipping. \citet{hanocka2019meshcnn} reports that these specific augmentations are most effective for deep learning mesh analysis, and therefore we use these augmentations as a starting point. Anisotropic scaling scales each vertex in a face at differing degrees, such that $\langle v_1, v_2, v_3 \rangle \in F$ becomes $\langle Sv_1, Sv_2, Sv_3 \rangle$, where each S is a coefficient sampled from a normal distribution. This heavily augments the 5-dimensional input feature, as the angles and edge-length ratios are changed. Vertex shifting rearranges certain vertices to different locations on the mesh, training the network to be invariant under vertex permutation. Edge flipping is applied for edges between two adjacent faces. The probabilities and scaling values of these augmentations are stochastically adjusted throughout training.

We note that, relative to image augmentations, the mesh augmentations are not as evident or distorting. They are even challenging to notice visually in a static figure, as only an interactive 3D viewing application would make the augmentations visible (hence why they are not provided in this paper). 
Nevertheless, we aim to propose our own strong augmentation policy compatible with contrastive learning that can show promising results.

As per \citet{chen2020simple}, we use NT-Xent as our contrastive loss. Let $\text{sim}\left(u,v\right)$ represent the cosine similarity between vectors $u$ and $v$. For outputted latent representations $\hat{x}_{i}$ and $\hat{x}_{j}'$, the contrastive loss function $\mathcal{L}$ is defined as 
\begin{equation}
    \mathcal{L}(\hat{x}_{i}, \hat{x}_{j}') = -\log\frac{\exp\left(\text{sim}\left(\hat{x}_i, \hat{x}_j'\right)/\tau\right)}{\sum^{2M_1}_{k=1}\mathbbm{1}_{[k\neq{i}]}\exp\left(\text{sim}\left(\hat{x}_i, \hat{x}_k'\right)/\tau\right)},
    \label{egn:loss}
\end{equation}
where $\mathbbm{1}_{[k\neq{i}]} \in \{0, 1 \}$ is an indicator function evaluating to 1 iff $k\neq{i}$ and $\tau$ is the temperature parameter.

Once contrastive learning is complete, we transfer the pre-trained encoder to a Mesh-UNet by attaching the encoder to a randomly-initialized decoder. Our Mesh-UNet is trained on a limited quantity of meshes $x$ and labels $y$ to predict segmented meshes $\hat{y}$. We perform traditional semantic segmentation with the pre-trained network using standard cross-entropy loss. Algorithm \ref{alg:train} formally details our entire training procedure.

\begin{algorithm}[t]
\caption{SSL-MeshCNN Mini-Batch Training}
\label{alg:train}
\begin{algorithmic}
\REQUIRE 
\STATE Training set of all meshes $x$ forming labeled dataset $\mathcal{D}_u$
\smallskip
\STATE Training set of meshes $x$ and corresponding segmentation labels $y$ forming dataset $\mathcal{D}_l$
\smallskip
\STATE Network architecture $\mathcal{F}_\theta=\left(\mathcal{F}_{1,\theta}, \mathcal{F}_{2,\theta}\right)$ with learnable parameters $\theta$; $\mathcal{F}_{1,\theta}$ is for SimCLR, $\mathcal{F}_{2,\theta}$ is for Mesh-UNet
\smallskip
\STATE Stochastic augmentation function $\text{Aug}\left(\cdot\right)$
\smallskip
\STATE Minibatch size $M_1, M_2$
\smallskip
\STATE Number of epochs for training $N_1, N_2$ 
\medskip
\smallskip
\REPEAT

\STATE  Sample minibatch of size $M_1$; each $x\sim \mathcal{D}_u$.
\smallskip
\STATE Compute latent representations by applying stochastic augmentations; $\hat{x}_{i} \leftarrow \mathcal F_{1,\theta}\left(\text{Aug}(x)\right), \hat{x}_{j}'\leftarrow \mathcal{F}_{1,\theta}\left(\text{Aug}(x)'\right)$ for $i,j\in\{1,\cdots, M_1\}$
\smallskip
\STATE Calculate a contrastive loss $\mathcal{L}\left(\hat{x}_{i}, \hat{x}_{j}' \right)$ with the equation \eqref{egn:loss} when $i=j$
\smallskip
\STATE Update $\mathcal{F}_{1,\theta}$ by back-propagating the loss gradient $\nabla_\mathcal{L}$
\UNTIL{$N_1$}
\medskip
\REPEAT
\STATE Sample minibatch of size $M_2$; each $x\sim \mathcal{D}_l$.
\smallskip
\STATE Compute segmentation predictions: $\hat{y}_i \leftarrow \mathcal F_{2,\theta}\left(x_i\right)$
\smallskip
\STATE Calculate cross-entropy loss $L(y_i,\hat{y}_i)$ for reference $y_i$ and prediction $\hat{y}_i$
\smallskip
\STATE Update $\mathcal{F}_{2,\theta}$ by back-propagating the loss gradient $\nabla_L$
\UNTIL{$N_2$}
\end{algorithmic}
\end{algorithm}

\section{Experimental Evaluation}

\subsection{Data and Training Setup}

\textbf{Data:} For experimentation, we use the Human Body Segmentation dataset introduced by \citet{maron2017convolutional}, which contains 381 meshes for training and 18 for testing. Each mesh comes with a semantic segmentation label containing 8 classes. For contrastive learning, we utilize the entire training set without the segmentation labels. For downstream segmentation, we use varying quantities of samples.

\textbf{Training:} For contrastive pre-training, we utilize all 381 meshes for training. For downstream segmentation, we conduct experiments by randomly sampling varying proportions of data: 5\%, 10\%, 25\%, 33\%, 50\%, 67\%, 75\%, and 100\%. For example, when using 10\% of the data, we train on only 38 randomly selected meshes and their corresponding segmentation labels. All segmentation experiments are performed with a standard Mesh-UNet. Each experiment is repeated 3 times and the means are reported. We implemented the models using PyTorch and trained using an NVIDIA GeForce GTX 1080.

\textbf{Hyperparameters:} We perform pre-training with minibatch size $M_1$ = 32 and downstream training with $M_2$ = 12. We pre-train for epochs $N_1$ = 100 and downstream train for epochs $N_2$ = 30. These training hyperparameters were experimentally tuned. We use the Adam optimizer with a learning rate of 0.0002 and group norm of 16. For our loss, we set $\tau$, the temperature parameter, to 0.7. The optimizer and loss hyperparameters were chosen based on previous works.

\textbf{Augmentations}: For anisotropic scaling, we randomly sample each scaling value from a normal distribution with $\mu = 1.0$ and $\sigma$ = 0.1. We perform vertex shifting with a probability of 0.2 and edge flipping with a probability of 0.05. 

\textbf{Evaluation:} Segmentation performance was evaluated using global accuracy. Our segmentation labels assign each edge to a class, so the accuracy is the percentage of edges which are correctly labeled by the network.

\subsection{Results and Discussion}

\begin{table}
    \setlength{\tabcolsep}{4pt}
    \centering
    \resizebox{\linewidth}{!}{
    \begin{tabular}{@{} c c ccccc@{}}
        \toprule
        \multirow{2}{*}{Dataset Proportion (\%)} & \phantom{a} & \multicolumn{5}{c}{Segmentation Accuracy (\%)} \\
        \cmidrule{3-7}
        &&  w/o SSL && w/ SSL && Diff \\
        \midrule
        5 &&  58.02 && 63.15 && 5.15 \\
        10 &&  63.89 && 66.29 && 2.40 \\
        25 &&  81.36 && 83.61 && 2.25 \\
        33 &&  83.98 && 85.08 && 1.10 \\
        50  && 85.25 && 87.84 && 2.59 \\
        67  && 86.69 && 88.97 && 2.28 \\
        75  && 87.52 && 90.35 && 2.83 \\
        100  && 88.58 && 90.50 && 1.92 \\
        \bottomrule
        \end{tabular}
    }
    \caption{Segmentation accuracy with and without contrastive learning pre-training at varying quantities of labeled data.}
    \label{tab:results}
\end{table}

\begin{figure}
    \centering
    \resizebox{\linewidth}{!}{
    \includegraphics[width = \linewidth]{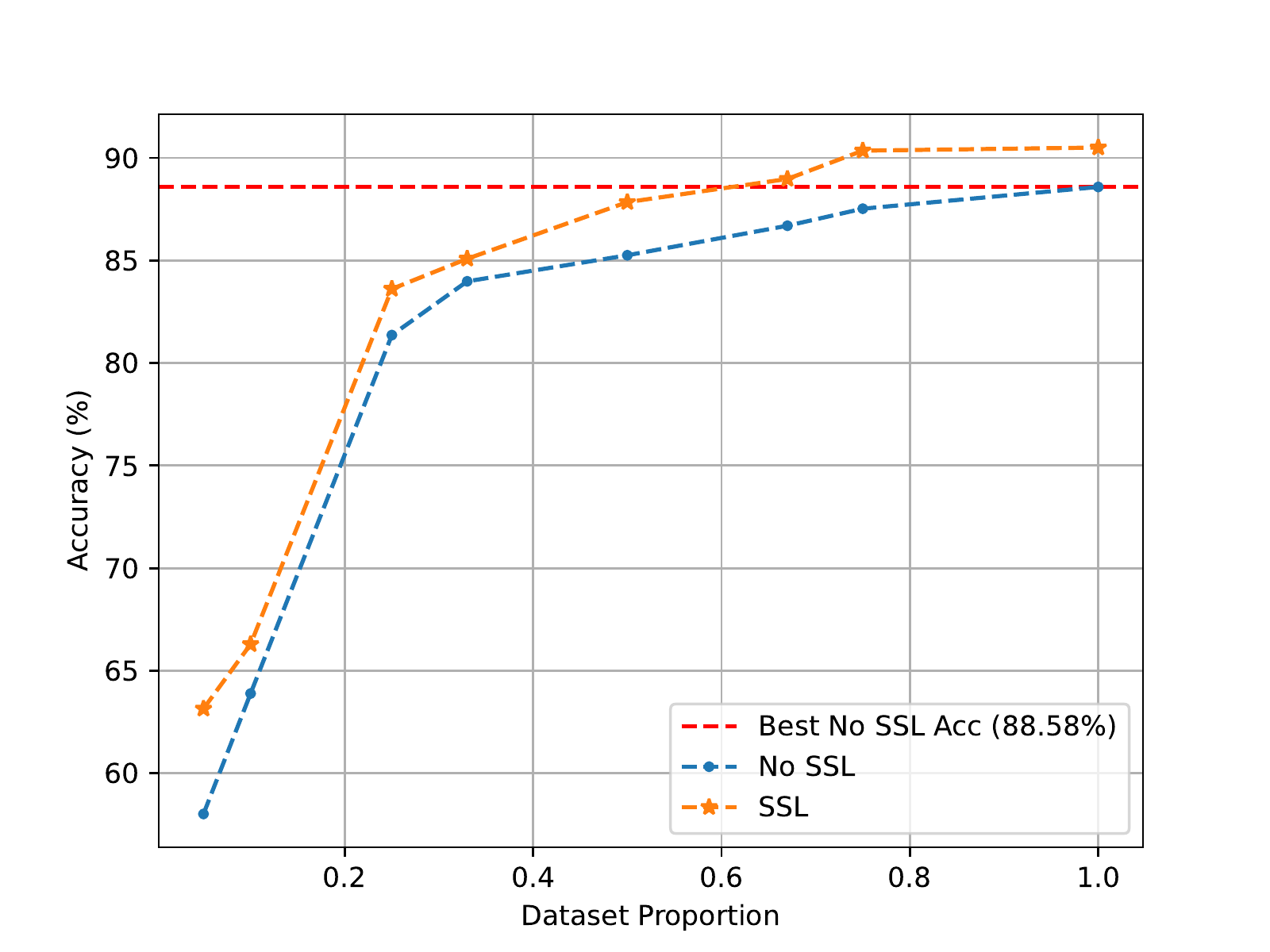}
    }
    \caption{Accuracy at varying quantities of labeled data with and without contrastive learning. Models trained with SSL have star points. Dashed red line indicates highest performance achieved by no SSL training.}
    \label{fig:accuracy}
\end{figure}

Table \ref{tab:results} and Figure \ref{fig:accuracy} display the segmentation accuracy at varying proportions of labeled data with and without self-supervised contrastive learning. Our results demonstrate that our constrastive learning method improves performance over no SSL baselines at all quantities of labeled data. At the lowest levels of supervision (5\% and 10\%), we report a five percentage point increase in accuracy as a result of SSL. The dashed red line in Figure \ref{fig:accuracy} marks the highest performance achieved without SSL, revealing that contrastive pre-training enables the network to match the fully-supervised performance when trained on just 67\% of the dataset. We can thus conclude that our preliminary contrastive learning framework reduces the need for supervision by 33\%, or that when building mesh datasets, only 67\% of the data must be manually labeled. Since we randomly selected samples for each experiment, we can conclude that when deciding which samples to label in a dataset, any random portion of samples can be labeled to achieve our result.

\begin{figure}
\centering
 \resizebox{\linewidth}{!}{%
  \begin{tabular}{ccccccc}
    \\
    \noalign{\smallskip}
    &
    {\Large 25} & {\Large 50} & {\Large 75}\\
    \cmidrule{2-4}
    {\Large w/o SSL}
    &
    \includegraphics[width=0.3\linewidth, trim={5cm 2.8cm 7.5cm 4.1cm},clip]{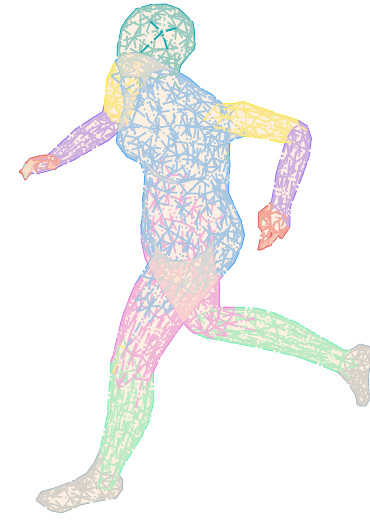}
    &
    \includegraphics[width=0.3\linewidth, trim={5cm 2.8cm 7.5cm 4.1cm},clip]{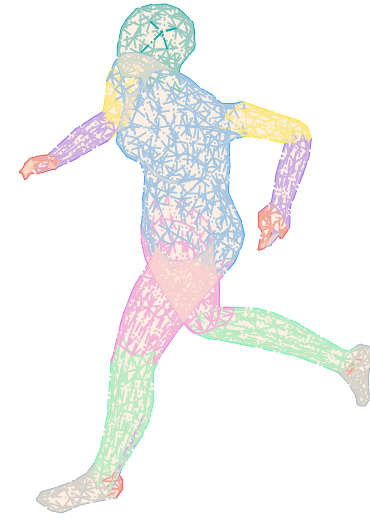}
    &
    \includegraphics[width=0.3\linewidth, trim={5cm 2.8cm 7.5cm 4.1cm},clip]{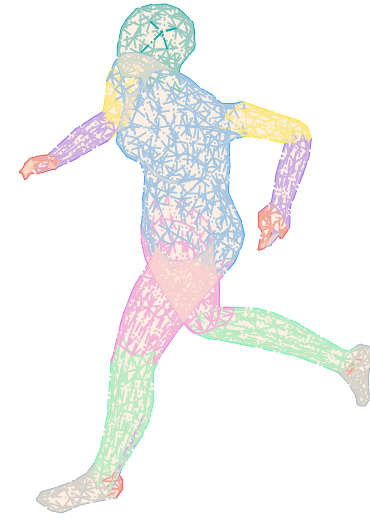}
    \\
    \noalign{\smallskip}
    {\Large w/ SSL} 
    &
    \includegraphics[width=0.3\linewidth, trim={5cm 2.8cm 7.5cm 4.1cm},clip]{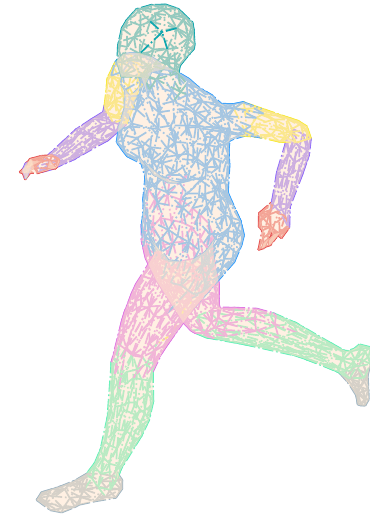}
    &
    \includegraphics[width=0.3\linewidth, trim={5cm 2.8cm 7.5cm 4.1cm},clip]{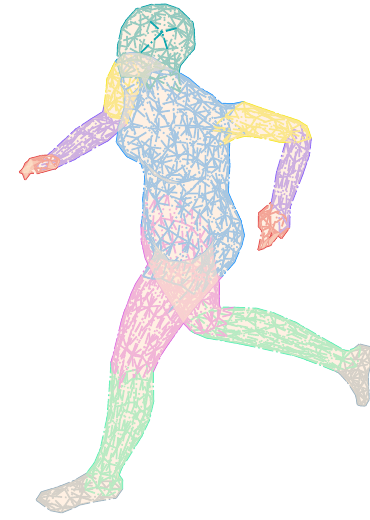}
    &
    \includegraphics[width=0.3\linewidth, trim={5cm 2.8cm 7.5cm 4.1cm},clip]{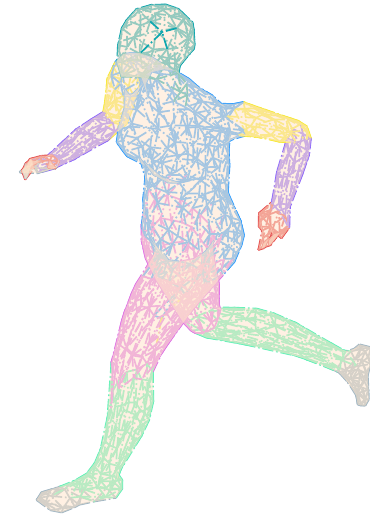}
    \\
    \end{tabular} 
    }
    \caption{Predicted segmentation visualizations at varying proportions of the dataset (\%) with and without SSL pre-training.}
    \label{fig:seg_vis}
\end{figure}

Figure \ref{fig:seg_vis} displays visualized segmentation results of our network at varying quantities of labeled data with and without contrastive pre-training. As shown, when pre-trained with self-supervised learning, the network achieves more accurate and representative segmentations than the fully-supervised baselines. The intersections between classes are more distinct and do not bleed into each other (fingers, knee, foot, etc.) with contrastive pre-training. The accuracies of these examples further confirm the superiority of our self-supervised pre-training.

\begin{figure}
    \centering
    \resizebox{\linewidth}{!}{
    \includegraphics[width = \linewidth]{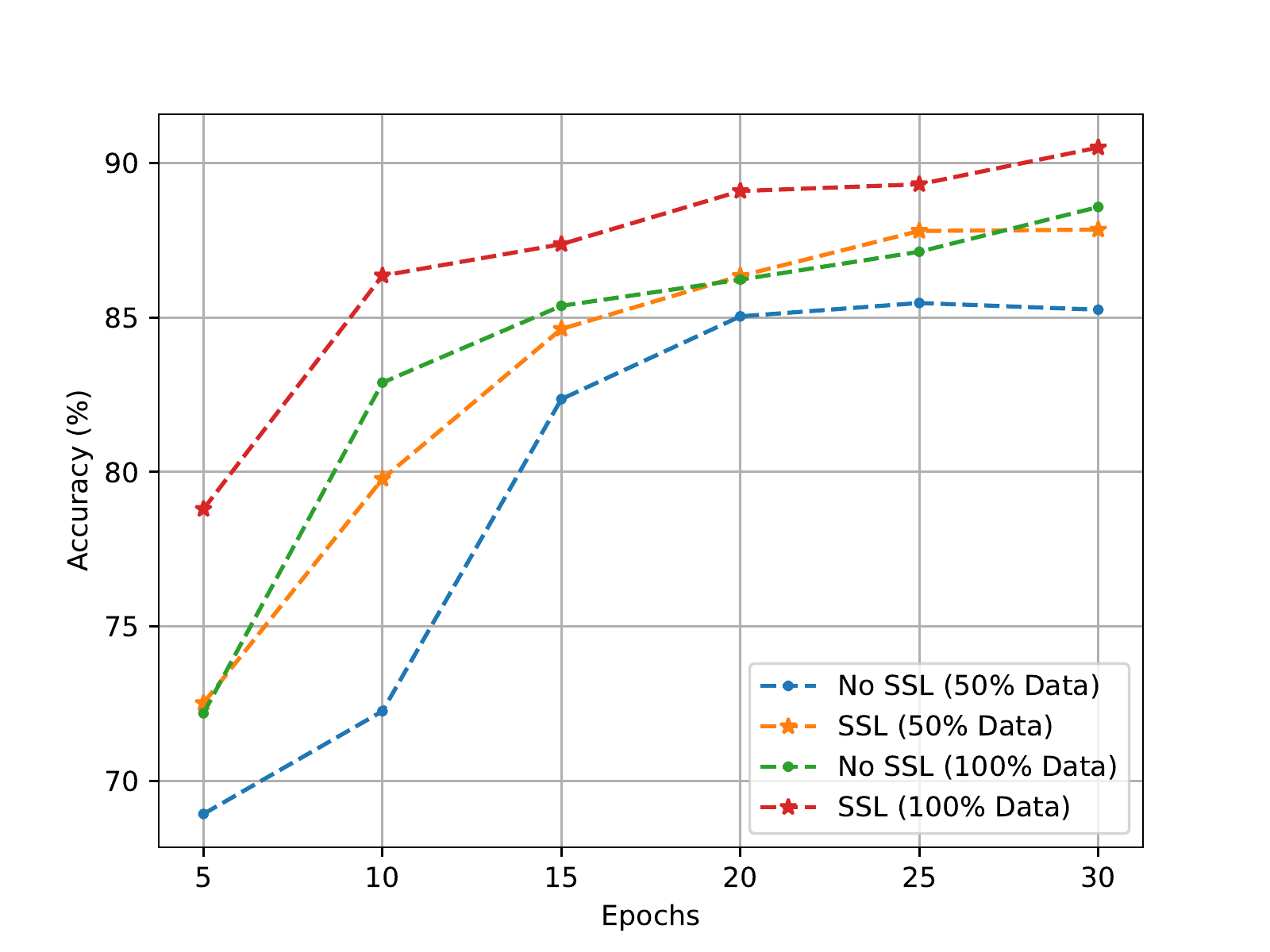}
    }
    \caption{Convergence rates (accuracy vs epochs) for training at 50\% and 100\% data proportions with and without SSL. Models trained with SSL have star points.}
    \label{fig:convergence}
\end{figure}

Figure \ref{fig:convergence} displays the accuracy at different epochs during training. As shown, when using SSL pre-training, the network achieves higher accuracies at earlier epochs and converges quicker. Moreover, we can see that with just 50\% of the data for downstream segmentation, using SSL can almost match the accuracy of 100\% of the dataset with no SSL. Overall, contrastive pre-training enables the network to converge faster and to higher accuracies than with no pre-training.

\begin{figure}
    \centering
    \resizebox{\linewidth}{!}{
    \includegraphics[width = \linewidth]{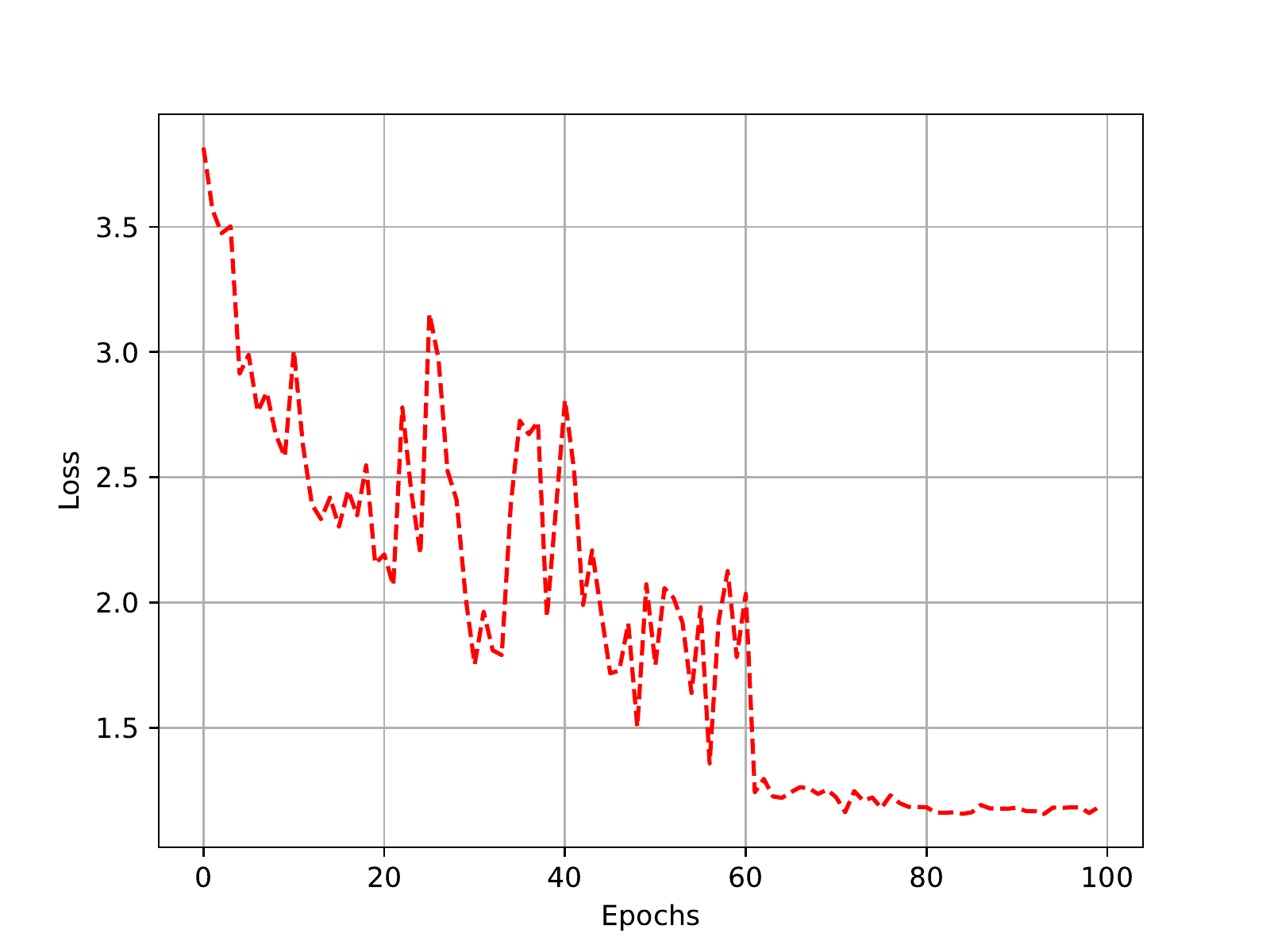}
    }
    \caption{Loss (NT-Xent) values at different epochs for contrastive pre-training.}
    \label{fig:losses}
\end{figure}

Figure \ref{fig:losses} display the loss throughout contrastive pre-training. After approximately 60 epochs, the loss decreases and stabilizes, implying that after a long training period, the network will converge and not learn better representations.

\section{Conclusions}

We have presented SSL-MeshCNN, a novel contrastive learning method designed for 3D meshes. We propose a tailored augmentation policy for the irregular 3D data format and pre-train the network to perform segmentation with reduced supervision. Our preliminary results confirm the effectiveness of our method at learning efficient representations, which enables the network to perform segmentation with less labeled meshes. We find that our contrastive pre-training can reduce the need for labeled examples for mesh segmentation by at least 33\%. Our future work will focus on designing more rigorous augmentation policies, constructing specialized model architectures, and experimenting with different datasets.

\bibliography{main}

\end{document}